\documentclass{llncs}

\usepackage{times}
\usepackage{epsfig}
\usepackage{graphicx}
\usepackage{amsmath}
\usepackage{amssymb}
\usepackage[linesnumbered,noend, noline, ruled]{algorithm2e}
\usepackage{subfigure} 
\usepackage{llncsdoc}
\usepackage[font=small,skip=-12pt]{caption}

\begin{document}

\title{Small Sample Learning of Superpixel Classifiers for EM Segmentation- Extended Version}

\author{Toufiq Parag, Stephen Plaza, Louis Scheffer\\ Janelia Farm Research Campus- HHMI, Ashburn, VA, USA.\\ \email{paragt@janelia.hhmi.org}}
\institute{}


\maketitle
\setlength{\textfloatsep}{10pt}
\setlength{\intextsep}{10pt}

\begin{abstract}
Pixel and superpixel classifiers have become essential tools for EM segmentation algorithms. Training these classifiers remains a major bottleneck primarily due to the requirement of completely  annotating the dataset which is tedious, error-prone and costly. In this paper, we propose an interactive learning scheme for the superpixel classifier for EM segmentation. Our algorithm is \lq active semi-supervised\rq~ because it requests the labels of a small number of examples from user and applies label propagation technique to generate these queries. Using only a small set ($<20\%$) of all datapoints, the proposed algorithm consistently generates a classifier almost as accurate as that estimated from a complete groundtruth. We provide segmentation results on multiple datasets to show the strength of these classifiers.

\end{abstract}

\section{Introduction}

Connectomics is an emerging field in neuroscience where the goal is to discern neural connectivity in an organism.  Recent advances of Electron Microscopy (EM) techniques have enabled us to image neurons and their components in an unprecedented level of details. The sizes of such datasets suggest that (semi-)automated region labeling or segmentation is the most viable strategy to conduct subsequent biological analysis. The outputs of such automated algorithms require manual correction afterwards~\cite{takemura13}.


Motivated by the advances in natural image segmentation techniques (see~\cite{arbelaez11} and references therein), there have been many fruitful attempts to segment neural regions recently ~\cite{funke12}\cite{kaynig10}\cite{vitaladevuni10} \cite{chklovskii10}\cite{jain11}\cite{andres08}\cite{andres12}. Most of these studies initially apply a pixel (2D or 3D)-wise classifier~\cite{jain10cvpr} to compute the boundary confidence at any location and produce an initial (over-)segmentation comprising superpixels through methods such as Watershed~\cite{meyer93}. Different approaches use different methods to refine, as well as register in anisotropic problem, the initial region labeling in order to generate the final segmentation. We adopt an Agglomerative or Hierarchical clustering scheme~\cite{chklovskii10}\cite{jain11} due to its advantages e.g., low space, time complexity and flexibility to tune for over/under segmentation.

Identifying a potential merge between two superpixels (either in 2D or 3D) through classification is a crucial step for almost all successful EM segmentation algorithms~\cite{funke12} \cite{kaynig10}\cite{vitaladevuni10}\cite{jain11}\cite{andres08}\cite{andres12}. This classifier may  act as a region boundary predictor;  given two adjacent superpixels, it classifies the separating boundary to be a true cell boundary or a false separation generated by the over-segmentation algorithm~\cite{andres12}\cite{jain11}\cite{nunez13}. For anisotropic dataset, the classifier may also be trained to identify which superpixels on different planes indeed  belong to the same neural body~\cite{funke12}\cite{kaynig10}\cite{vitaladevuni10}. This study investigates an interactive small sample learning method to generate a robust and accurate classifier to be used primarily for region boundary detection. 

\begin{figure}[t]
\begin{center}
\subfigure[]{\includegraphics[width=0.24\columnwidth, height=0.21\columnwidth]{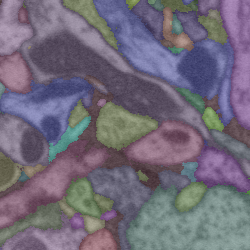}\label{F:COMPLETE_GROUNDTRUTH}}
~~~~\subfigure[]{\includegraphics[width=0.22\columnwidth, height=0.21\columnwidth]{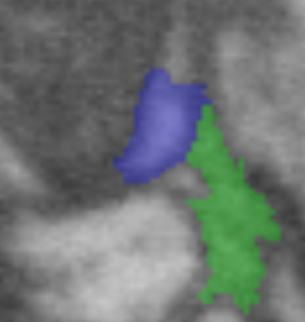}\label{F:INTERACTIVE-1}}
\subfigure[]{\includegraphics[width=0.22\columnwidth, height=0.21\columnwidth]{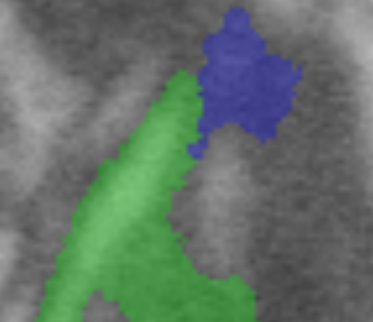}\label{F:INTERACTIVE-2}}
\subfigure[]{\includegraphics[width=0.22\columnwidth, height=0.21\columnwidth]{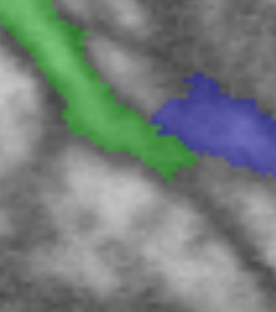}\label{F:INTERACTIVE-3}}
\end{center}
\caption{\scriptsize Leftmost: complete groundtruth. Right 3 images: examples of queries generated by the proposed algorithm. }\label{F:INTRO}
\end{figure}

While it is critical to have highly accurate classifiers for accurate segmentation~\cite{andres08} \cite{jain11},  training such classifiers, both pixel and superpixelwise, is considered to be a major bottleneck in EM segmentation literature~\cite{helmstaedter13}. On one hand, training algorithms typically demand complete groundtruth labels, which assigns a label to each location of a volume or a set of images as displayed in Figure~\ref{F:INTRO}\subref{F:COMPLETE_GROUNDTRUTH} where each color represents a label. Generating such labeling from either the actual grayscale images or a preliminary segmentation output is costly, tedious and entails the risk of human errors (e.g., due to loss of attention over time) and therefore could significantly stifle the performance of such predictors~\cite{helmstaedter13}. While dependence on complete groundtruth may sometimes be alleviated by using the interactive tool Ilastik~\cite{ilastik11} for pixel-wise detection~\cite{nunez13}, it can not be eliminated altogether due to the subsequent superpixel classifier training step. The work of~\cite{andres12} attempts to train the superpixel boundary predictor in an interactive fashion similar to Ilastik, but it has not been shown to be comparable, in terms of accuracy, to the one learned from a complete labeling.

On the other hand, the frameworks that require multiple training phases~\cite{jain11}\cite{nunez13}  are less amenable to have a small sample variant. Since  a single class, such as cytoplasm, may comprise several sub-classes possessing widely varying characteristics, it seems to be crucial for these algorithms to recursively refine the hypotheses and accumulate training sets for satisfactory performance. In contrast, a context-aware approach~\cite{supple14context}, which processes different components (sub-structures) of the same class separately, was demonstrated to attain improved segmentation performance with predictors trained from fewer examples than those in~\cite{jain11}\cite{nunez13}. The context-aware approach of~\cite{supple14context} motivated us to pursue a training method for a superpixel boundary classifier with even fewer numbers of examples while retaining the accuracy of that learned on complete groundtruth. 

{\it This paper  proposes an active semi-supervised algorithm for training a superpixel face detector utilizing a significantly small subset ($< 20\%$) of all such faces} .  In the proposed framework, the user is repeatedly asked to assign binary class labels (true/false) to a few of all region boundaries, such as the ones shown in Figure~\ref{F:INTRO}\subref{F:INTERACTIVE-1}, \subref{F:INTERACTIVE-2}, \subref{F:INTERACTIVE-3}, generated by an over-segmentation algorithm\footnote{\scriptsize Access to an over-segmented volume computed by an interactive (or other limited sample) pixel-wise classifier and an appropriate region growing algorithm is assumed in this study.}.  Provided these labels, the algorithm updates the classifier learned so far and generates a new set of queries for the next round.  

Active learning algorithm are known to suffer from inconsistent performance and low noise tolerance~\cite{iwal09}. Experimentally, the incorporation of a semi-supervised method has been shown to achieve a greater degree of robustness for \lq actively\rq~ trained predictors~\cite{zhu03} \cite{muslea02icml}. Similar to~\cite{zhu03}, our proposed interactive learning employs a semi-supervised method along with a classifier to identify the most useful examples to be queried next.  {\it The boundary predictors produced by the proposed method are shown to consistently achieve  (almost) the same performance as that learned with full groundtruth} and to be more robust than those trained using several standard active learning techniques.

Our framework has a substantial implication on the overall EM reconstruction process. Together with an interactive pixel boundary detector (e.g., Ilastik), the proposed method paves the way for EM segmentation without exhaustive annotation. This enables us to learn and apply segmentation quickly and is advantageous for  large volume reconstruction (e.g., whole animal brain) where one may anticipate to learn different predictors for different areas to improve accuracy. A quick segmentation  output on preliminary images, generated with different sample preparations, could also assist the imaging expert to decide the optimal preparation during EM imaging.



\section{Interactive Learning of Region Boundary Predictor}

Let us suppose the initial over-segmentation process generated $N$ superpixels  ${\cal S} = \{ S_1, S_2, \dots, S_N\}$ on an EM dataset with $M$ neurites (neuronal regions) where $ N \gg M$. Let $L(S)$ be the neurite region that $S$ actually belongs to. Our goal is to iteratively merge these $N$ superpixels such that each $S_i, i = 1, 2, \dots, N$ is merged into its corresponding $L(S_i)$.

We denote a boundary between two oversegmented regions by a pair of regions $e \triangleq \{ S_i, S_j\}$ and the set of all such boundaries by $E$. In a graph representation, each of the regions $S_i$ is considered to be a node and the boundary or face between two regions is regarded as an edge -- a notation we will be using throughout the paper. Also, let the binary boundary label map $B: {\cal S} \times {\cal S} \to \{-1, 1\}$ assign a 1 to a boundary that actually separates one neurite region from another and a -1 to the boundary incorrectly generated due to over-segmentation.  In agglomerative clustering methods, a real-valued superpixel boundary confidence function $h :{\cal S} \times {\cal S} \to \mathbb{R}$ approximates $B(e)$. In this paper, we describe how this predictor $h$ can be estimated from a small subset rather than all boundaries of $E$.


\subsection{Active Semi-supervised learning}
The primary challenge of an active learning algorithm is to fit a classifier with few examples respecting the actual class boundaries in the feature distribution of the whole dataset. Without any additional information, any classifier will concentrate on separating the members of different classes within the small sample set at hand. It is, however, possible to extrapolate the labels of other datapoints, through the most similar ones to the small set, by  a semi-supervised algorithm called the label propagation technique~\cite{semi-super06}\cite{zhu03}. A datapoint is more informative for classification if it is classified differently than the examples \lq nearby\rq~(or the ones it is strongly connected to based on the affinities among them) -- in other words, the samples for which the generative label propagation estimate disagrees with the discriminative prediction.

The proposed  algorithm starts with a small subset $E_l \subset E$ which is initially fully labeled. A (Random Forest~\cite{breiman01}) classifier is trained on this initial dataset $E_l$ and the confidences of all the remaining edges $E_u = E \setminus E_l$ are computed. In addition, another set of confidences for $E_u$ is computed by a generative model. The disagreement among these two types of estimates are quantified in a ranking formula. The first $k$ examples in descending order of disagreement measure are presented to the user as queries. The set $E_l$ is augmented by this new queries and the whole process is repeated until some stopping criterion is satisfied.

\noindent \textbf{\textit{Generative view:}} The generative view for our approach consists of a graph based label propagation technique of semi-supervised learning~\cite{semi-super06}\cite{zhu03}. Let us denote $\mathbf{x}$ to be the feature representation of a boundary sample $e$. Given a set $E$ of examples, this algorithm computes a pairwise affinity matrix $W = \exp\bigl \{ -{1 \over 2} (\mathbf{x}_i - \mathbf{x}_j)^T \Sigma^{-1} (\mathbf{x}_i - \mathbf{x}_j) \bigr \}$ where $e_i, e_j \in E$ and its corresponding degree $D$ and Laplacian matrix $L = D - W$. The smoothness on the labels $\mathbf{y}$, given the pairwise affinities, can be enforced by requiring the quantity ${1 \over 2} \sum_{i \sim j} W_{ij} (y_i - y_j)^2 = \mathbf{y}^T W \mathbf{y} $ to be minimized. This energy term is minimized at $(D-W) \mathbf{y} = 0$.

Provided a set of known labels $\mathbf{y}_l$ for $e \in E_l$, the factorization of the labels and weight matrix enables us to compute the unknown labels $\mathbf{y}_u$ for the remaining set $E_u$ as follows.
\begin{equation}\label{E:SEMI_SUPER}
{\small
\mathbf{y} = \left [ \begin{array}{cc}
\mathbf{y}_{l} \\
\mathbf{y}_{u}
\end{array} \right ],  ~~~~
W = \left [ \begin{array}{cc}
W_{ll} ~&~ W_{lu} \\
W_{ul} ~&~ W_{uu}
\end{array} \right ], ~~~~ L_{uu}~\mathbf{y}_{u} ~=~ W_{ul}~ \mathbf{y}_l, }
\end{equation}
where $L_{uu}$ is the corresponding graph Laplacian of $W_{uu}$. Relaxing the values of $\mathbf{y}_u$ to real values, this system of linear equations can be solved efficiently by existing algorithms. 

\noindent \textbf{\textit{Ranking:}}Given the real valued confidences $h_c(e),~ e \in E_u$ from the current classifier (discriminative view) and the estimates $\mathbf{y}_u$ of the label propagation method, we use the following formula to compute disagreement between them.

\begin{equation}
R(e) = 1 - h_c(e)~y_u(e).
\end{equation}

The $k$ boundaries with largest $R(e)$ constitute the set $E_q$ of queries. The sets of edges are updated as follows: $E_l = E_l \cup E_q, E_u = E_u \setminus E_q.$

\section{Implementation Details}\label{S:IMPLEMENTATION}

An efficient solver for the linear system in Equation~\ref{E:SEMI_SUPER} is essential for the implementation of an interactive learner. Notice that, the  Laplacian matrix $L_{uu}$ is a symmetric diagonally dominant one. Fortunately, it has been shown that such systems can be solved in nearly linear time in terms of the number of edges on the graph~\cite{spielman08}\cite{koutis11}. In our implementation, we employed an efficient Algebraic Multigrid solver freely available at~\cite{demidov}. 

Similarly, we need a classifier for the discriminative view to be trained quickly. One of the reasons for our choice of Random Forest classifier~\cite{breiman01} is that the decision trees within the forest can  be trained in parallel and therefore can be computed efficiently on multi-core machines.  

To reduce redundancy, the initial labeled set $E_l$ was populated by the $k$ centers of the output of a clustering algorithm (in our case k-means). Another alternative is to select the $k$ datapoints with the largest degrees (w.r.t $W$) which are not neighbors to one another. The latter approach is deterministic and faster than clustering algorithms, but performs with same degree of accuracy in our dataset.

\section{Experiments and Results}\label{S:RESULT}

The interactive training method has been tested for 3D FIBSEM volumes as well as 2D ssTEM images (without alignment across planes). We generate the over-segmentation from the output of a multi-class (e.g., cell boundary, cytoplasm, mitochondria,  mitochondria boundary) \emph{pixel classifier trained on only few pixels selected from the volume/images} using Ilastik~\cite{ilastik11} for both data modalities. Following~\cite{andres12}\cite{supple14context}, the boundary predictor $h_c$ is trained only on cytoplasm superpixel boundaries with the superpixel features similar to ~\cite{supple14context}\cite{andres08}. To compute the affinities of $W$, a diagonal $\Sigma$ was used where $\Sigma(a,a)$ is the variance of $a$-th feature.

This paper reports the segmentation performance of a context-aware agglomeration strategy~\cite{supple14context} given the predictor $h_c$ trained by the proposed method. In this agglomeration scheme, the cytoplasm superpixels are first clustered using the predictor $h_c$ and then the mitochondria are merged based on their boundary ratios. The performances were  measured using split-VI and split-RI values as described in~\cite{supple14context}. We also report the segmentation performance of the Global method~\cite{andres12} with $h_c$ learned by our method.

The code for this paper and for~\cite{supple14context} can be found at \url{https://github.com/janelia-flyem/NeuroProof.git}.

\subsection{FIBSEM data} 
Initial over-segmentation on two training volumes (Tr set1, TrSet2, each of size $250^3$) generated roughly 30,000 edges each, the proposed algorithm utilized $3\%$ of all edges to populate the initial $E_l$. At each iteration, $10$ new samples were queried until the total number of samples reaches $5000$ (roughly $17\%$) at which point the algorithm terminates. We also trained the following standard active learning techniques and compared the performances on two different $520^3$ volumes(Test vol 1, Test vol 2) : 1) Bootstrap variant of Importance Weighted Active Learning ~\cite{iwal09}. 2) Active version of Co-training method~\cite{zhu09book} where the initial set of $E_l$ is divided and two different Random Forest classifiers were learned. Each example, on which the predictions of these two classifiers differ, is queried and inserted on the training set of the one that misclassified it. 3)Uncertain queries:  train a Random forest from small set of samples and query all samples with prediction in range $[-0.3, 0.3]$. 4) Random queries. Unless otherwise specified, all the parameters for these methods were kept the same as those of the proposed method. 
\begin{figure}[t]
\begin{center}
\subfigure{
\includegraphics[width=0.32\columnwidth, height=0.28\columnwidth]{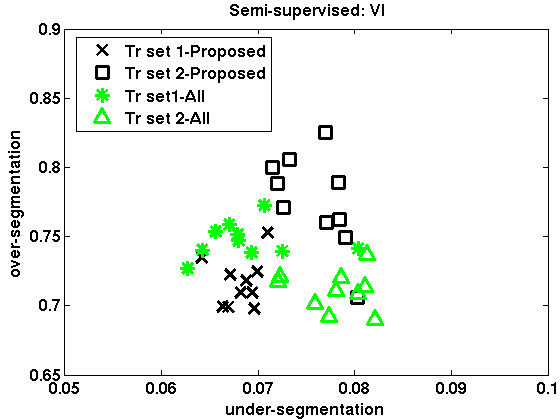}}
\subfigure{
\includegraphics[width=0.32\columnwidth, height=0.28\columnwidth]{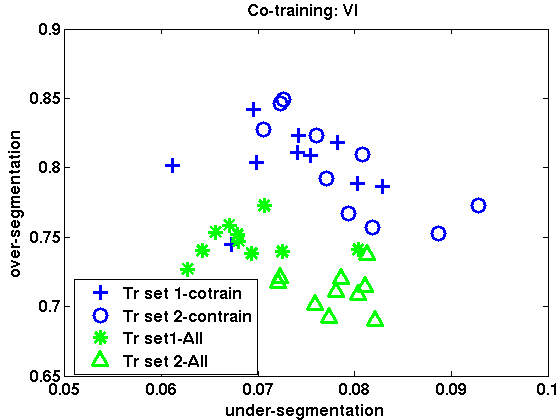}}
\subfigure{
\includegraphics[width=0.32\columnwidth, height=0.28\columnwidth]{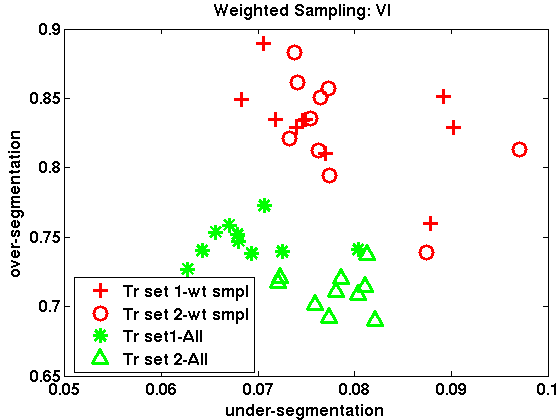}}
\setcounter{subfigure}{0}
\subfigure[Proposed]{
\includegraphics[width=0.32\columnwidth, height=0.28\columnwidth]{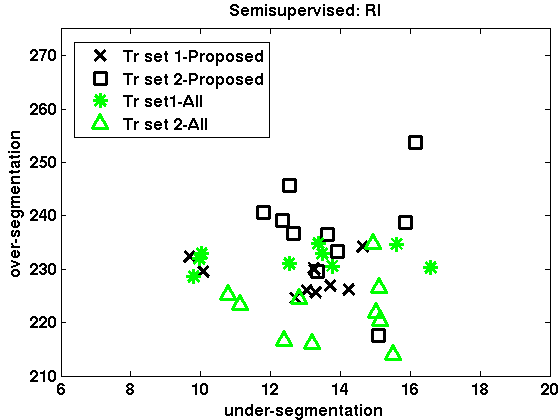}\label{F:RESULT_VI}}
\subfigure[Co-training]{
\includegraphics[width=0.32\columnwidth, height=0.28\columnwidth]{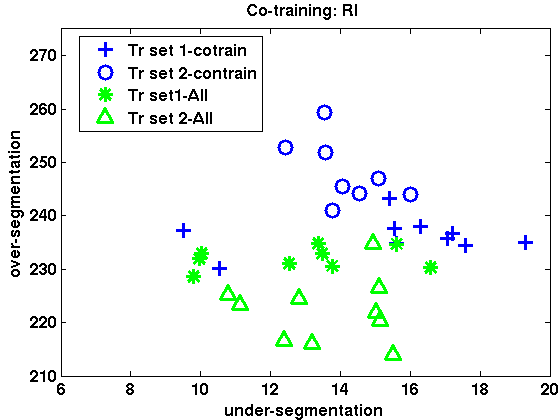}\label{F:RESULT_VI_COTRAIN}}
\subfigure[Importance Weighted~\cite{iwal09}]{
\includegraphics[width=0.32\columnwidth, height=0.28\columnwidth]{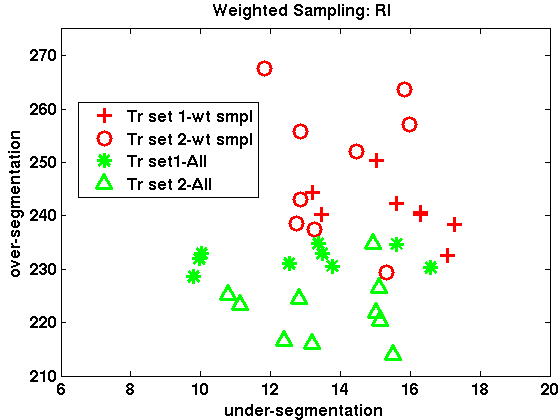}\label{F:RESULT_VI_IWAL}}
\end{center}
\caption{\scriptsize Segmentation error on FIBSEM Test volume 1.  Top: split-VI and bottom: split-RI.}
\label{F:RESULT_Q_FIB1}
\end{figure}

\begin{table}
\begin{center}
\caption{\scriptsize Split-VI average and standard deviation on FIBSEM Test vol 1}
\label{T:RESULT_VI_FIB1}
\begin{tabular}{ | c | c | c | c | c | }
  \hline
  Algorithm & \multicolumn{2}{|c|}{split-VI, Trn set 1}& \multicolumn{2}{|c|}{split-VI, Trn set 2}\\
  \hline
   & false merge & false split & false merge & false split \\
 \hline
 
  All  & $0.0688 \pm 0.005$ & $0.7469 \pm 0.0127$ & $0.0779 \pm 0.0035$ & $0.7113 \pm 0.0141$  \\
  Proposed  & $0.06814 \pm 0.002$ & $0.7167 \pm 0.0176$ & $0.0759 \pm 0.0032$ & $0.7758 \pm 0.0337$  \\
  co-train  & $0.0732 \pm 0.0065$ & $0.8027 \pm 0.0261$ & $0.07919 \pm 0.0071$ & $0.7998 \pm 0.0362$  \\
  iwal~\cite{iwal09} & $0.0778 \pm 0.0081$ & $0.832 \pm 0.0328$ & $0.0788 \pm 0.0075$ & $0.8268 \pm 0.0409$  \\
  uncertain & $0.0653 \pm 0.0042$ & $0.7538 \pm 0.0396$ & $0.0777 \pm 0.0048$ & $0.777 \pm 0.042$  \\
  random & $0.071 \pm 0.0032$ & $0.8208 \pm 0.0185$ & $0.0759 \pm 0.0071$ & $0.8032 \pm 0.0353$  \\
\hline
\end{tabular}
\end{center}
\end{table}

\begin{figure}[h]
\begin{center}
\subfigure{
\includegraphics[width=0.32\columnwidth, height=0.28\columnwidth]{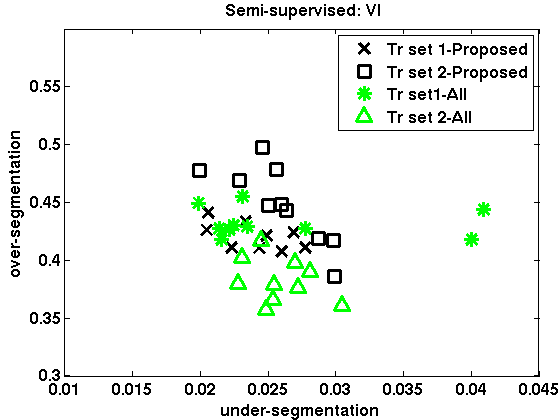}}
\subfigure{
\includegraphics[width=0.32\columnwidth, height=0.28\columnwidth]{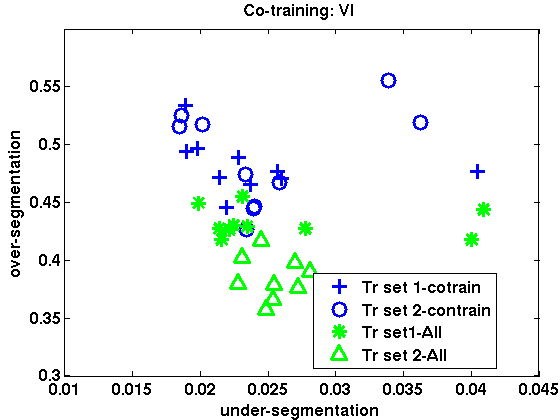}}
\subfigure{
\includegraphics[width=0.32\columnwidth, height=0.28\columnwidth]{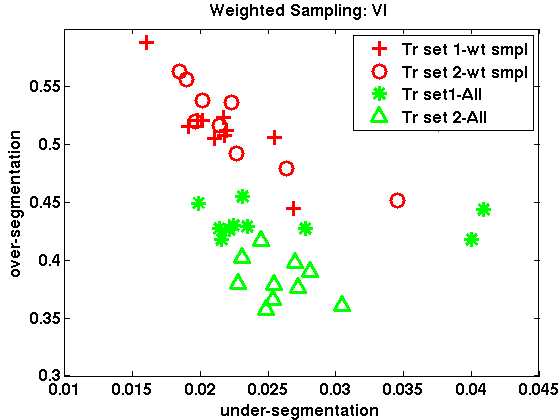}}
\setcounter{subfigure}{0}
\subfigure[Proposed]{
\includegraphics[width=0.32\columnwidth, height=0.28\columnwidth]{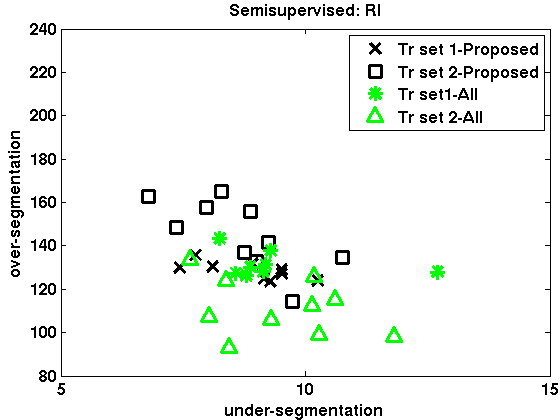}\label{F:RESULT_VI}}
\subfigure[Co-training]{
\includegraphics[width=0.32\columnwidth, height=0.28\columnwidth]{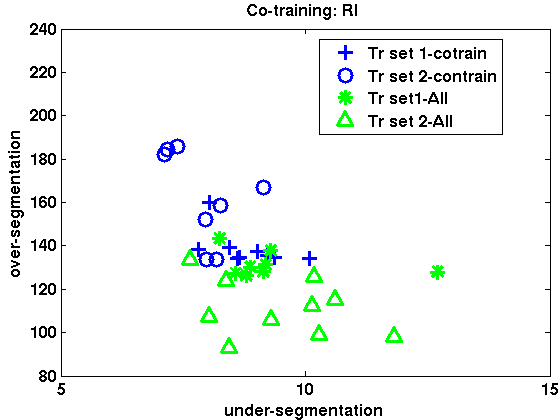}\label{F:RESULT_VI_COTRAIN}}
\subfigure[Importance-weighted~\cite{iwal09}]{
\includegraphics[width=0.32\columnwidth, height=0.28\columnwidth]{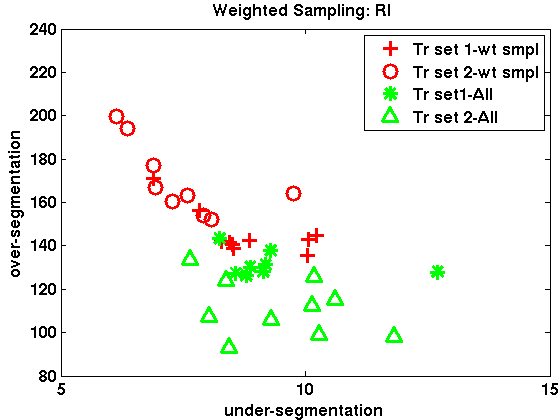}\label{F:RESULT_VI_IWAL}}
\end{center}
\caption{\scriptsize Segmentation error on FIBSEM Test volume 2.  Top: split-VI and bottom: split-RI.}
\label{F:RESULT_Q_FIB2}
\end{figure}

In order to evaluate robustness, the training algorithms are executed 10 times on two training volumes (Tr set 1 and Tr set 2) and the corresponding segmentation errors, in terms of split-VI and split-RI,  are displayed on Figure~\ref{F:RESULT_Q_FIB1} for  Test vol 1, and Figure~\ref{F:RESULT_Q_FIB2} for Test vol 2. The threshold $\delta_c$, which remains fixed for the multiple trials, was chosen such that the under-segmentation error of any particular actively trained predictor remains close to that produced by the one learned from full groundtruth agglomerated upto $\delta_c = 0.2$. Each plot compares the errors of the two predictors estimated by a certain active learning scheme (e.g., proposed in black + and square on Figure~\ref{F:RESULT_Q_FIB1}\subref{F:RESULT_VI}) with those of  boundary predictors learned with full groundtruth (Green * and triangle). \emph{For both the test volumes (Figures~\ref{F:RESULT_Q_FIB1} and~\ref{F:RESULT_Q_FIB2}), the over and under-segmentation errors of proposed method are the closest to those of full groundtruth training model}. The means and std deviation of split-VI are reported in Tables~\ref{T:RESULT_VI_FIB1} and~\ref{T:RESULT_VI_FIB2}. 

\begin{table}
\begin{center}
\caption{\scriptsize Split-VI average and standard deviation on FIBSEM Test vol 2}
\label{T:RESULT_VI_FIB2}
\begin{tabular}{ | c | c | c | c | c | }
  \hline
  Algorithm & \multicolumn{2}{|c|}{split-VI, Trn set 1}& \multicolumn{2}{|c|}{split-VI, Trn set 2}\\
  \hline
   & false merge & false split & false merge & false split \\
 \hline
 
  All  & $0.0262 \pm 0.0077$ & $0.4323 \pm 0.0127$ & $0.0258 \pm 0.0023$ & $0.3826 \pm 0.0192$  \\
  Proposed  & $0.0241 \pm 0.0024$ & $0.4206 \pm 0.0109$ & $0.0258 \pm 0.0031$ & $0.4484 \pm 0.0340$  \\
  co-train  & $0.0239 \pm 0.0063$ & $0.4819 \pm 0.0236$ & $0.0247 \pm 0.0059$ & $0.4892 \pm 0.0428$  \\
  iwal~\cite{iwal09} & $0.0213 \pm 0.003$ & $0.5145 \pm 0.0345$ & $0.0242 \pm 0.0066$ & $0.5174 \pm 0.0346$  \\
  uncertain & $0.0308 \pm 0.0191$ & $0.477 \pm 0.0357$ & $0.0285 \pm 0.003$ & $0.430 \pm 0.0536$  \\
  random & $0.0231 \pm 0.0084$ & $0.4953 \pm 0.0254$ & $0.0237 \pm 0.0068$ & $0.4815 \pm 0.0481$  \\
\hline
\end{tabular}
\end{center}
\end{table}

\begin{figure}[h]
\begin{center}
\subfigure{
\includegraphics[width=0.3\columnwidth, height=0.27\columnwidth]{anirban-proposed-vi} }
\subfigure{
\includegraphics[width=0.3\columnwidth, height=0.27\columnwidth]{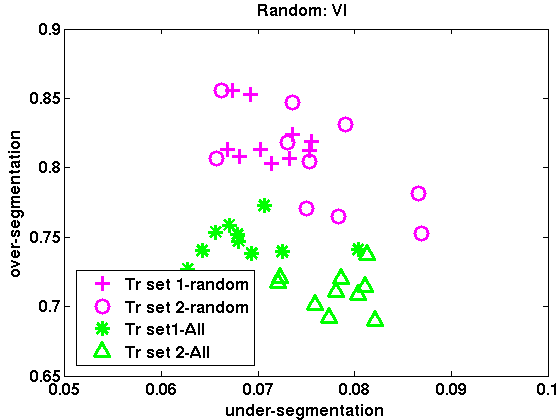} }
\subfigure{
\includegraphics[width=0.3\columnwidth, height=0.27\columnwidth]{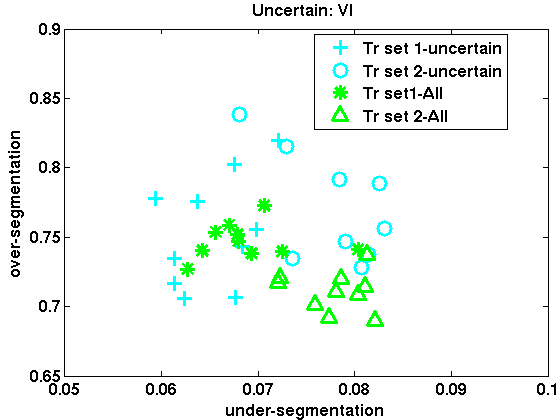} }
\setcounter{subfigure}{0}
\subfigure[\scriptsize Proposed]{
\includegraphics[width=0.3\columnwidth, height=0.27\columnwidth]{anirban-proposed-ri} }
\subfigure[\scriptsize Random]{
\includegraphics[width=0.3\columnwidth, height=0.27\columnwidth]{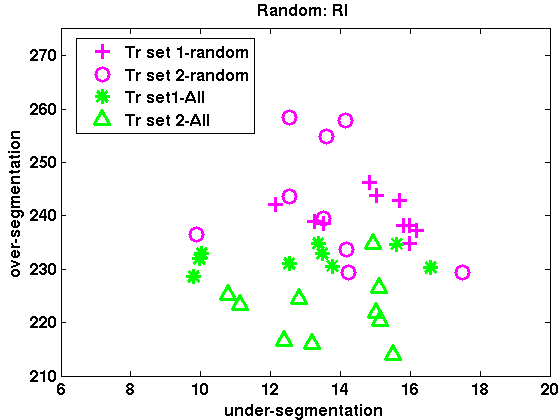} }
\subfigure[\scriptsize Uncertain]{
\includegraphics[width=0.3\columnwidth, height=0.27\columnwidth]{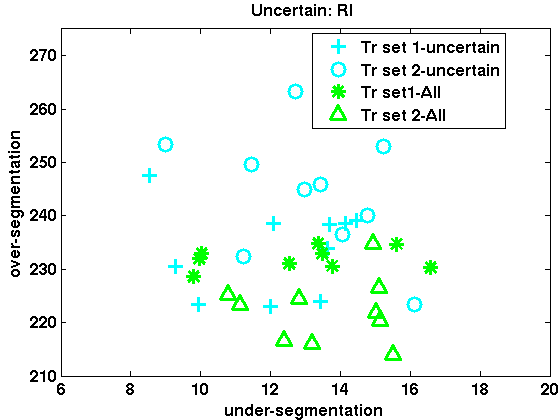} }
\end{center}
\caption{\scriptsize Segmentation performances on FIBSEM Test Vol 1. Top: split-VI and bottom: split-RI.}
\label{F:RESULT_VOL1_RANDOM}
\end{figure}
\begin{figure}[h]
\begin{center}
\subfigure{
\includegraphics[width=0.3\columnwidth, height=0.27\columnwidth]{fg-proposed-vi} }
\subfigure{
\includegraphics[width=0.3\columnwidth, height=0.27\columnwidth]{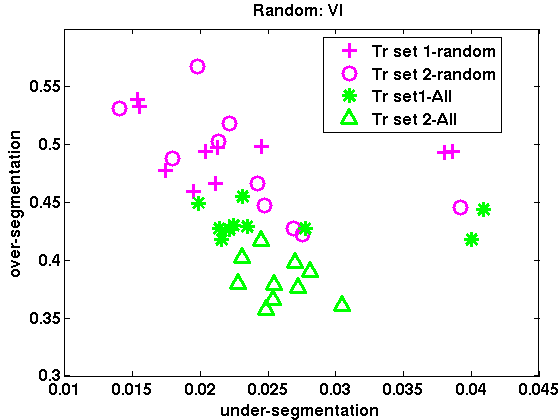} }
\subfigure{
\includegraphics[width=0.3\columnwidth, height=0.27\columnwidth]{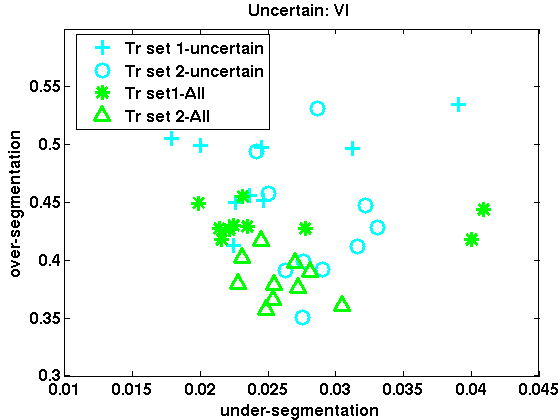} }
\setcounter{subfigure}{0}
\subfigure[\scriptsize Proposed]{
\includegraphics[width=0.3\columnwidth, height=0.27\columnwidth]{fg-proposed-ri} }
\subfigure[\scriptsize Random]{
\includegraphics[width=0.3\columnwidth, height=0.27\columnwidth]{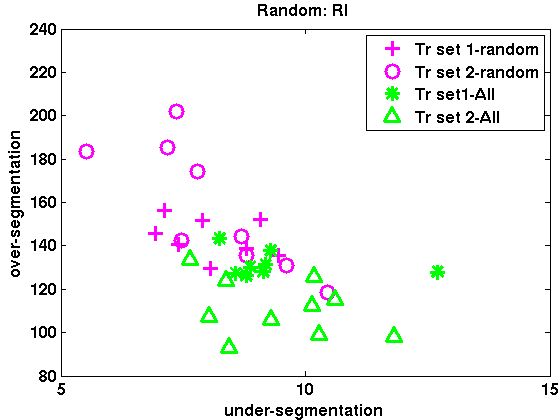} }
\subfigure[\scriptsize Uncertain]{
\includegraphics[width=0.3\columnwidth, height=0.27\columnwidth]{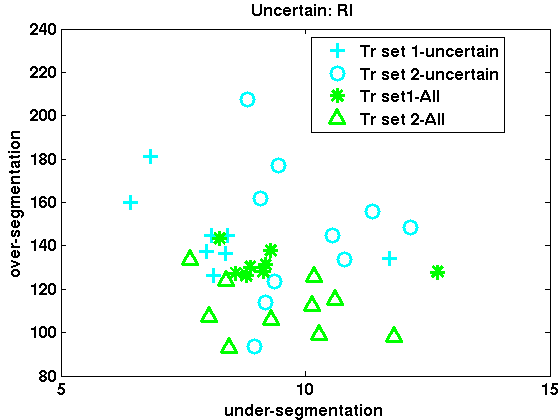} }
\end{center}
\caption{\scriptsize Segmentation performances on FIBSEM Test Vol 2. Top: split-VI and bottom: split-RI.}
\label{F:RESULT_VOL2_RANDOM}
\end{figure}

\begin{figure}[h]
\begin{center}
\subfigure[\scriptsize Query set error]{
\includegraphics[width=0.32\columnwidth, height=0.27\columnwidth]{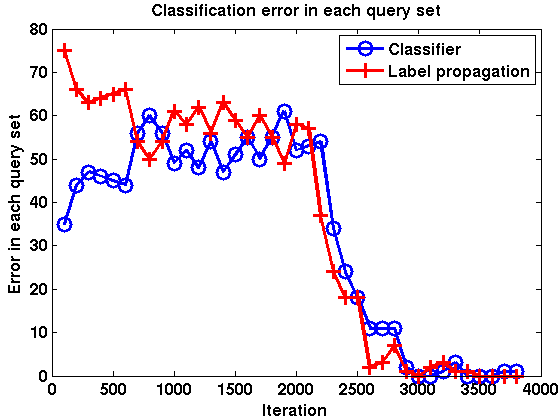}\label{F:ERROR_QUERY}}
\subfigure[\scriptsize Mutually exclusive error]{
\includegraphics[width=0.32\columnwidth, height=0.27\columnwidth]{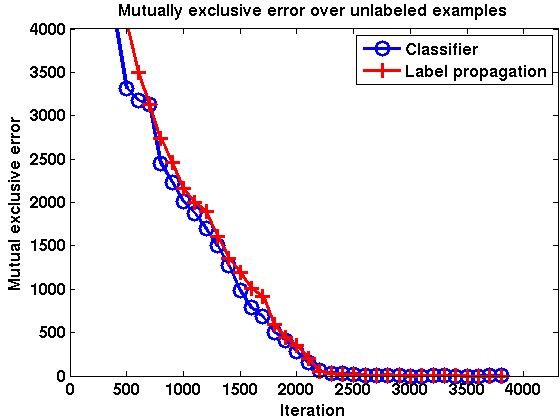}\label{F:ERROR_MUTUAL}}
\subfigure[\scriptsize Increase in classification accuracy]{
\includegraphics[width=0.32\columnwidth, height=0.27\columnwidth]{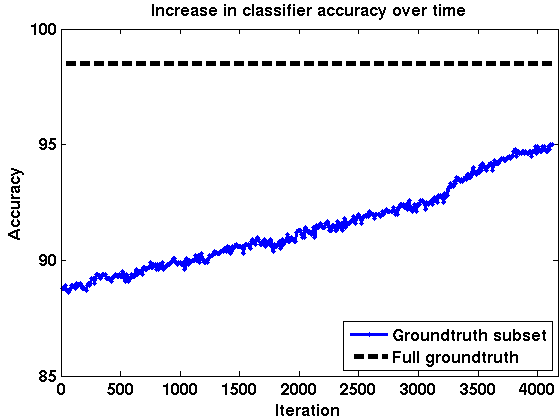}\label{F:ACCURACY}}
\end{center}
\caption{\scriptsize Analyzing errors of discriminative and generative views.}
\label{F:ANALYZE}
\end{figure}

For practical reasons, we are constrained to use far less samples than what is required by the importance weighted sampling method (Column~\subref{F:RESULT_VI_IWAL} of Figures~\ref{F:RESULT_Q_FIB1} and~\ref{F:RESULT_Q_FIB2} ) to achieve an accuracy similar to the full groundtruth classifier as suggested by the results in~\cite{iwal09}. The co-training algorithm (Column~\subref{F:RESULT_VI_COTRAIN} of Figures~\ref{F:RESULT_Q_FIB1} and~\ref{F:RESULT_Q_FIB2} ) is apparently less capable than the proposed strategy in reinforcing the performances of the two hypotheses -- this is not surprising since  each of these hypotheses is trained on half of the training set. The random and uncertain query models both manifests inconsistent results over multiple trials ( Figures~\ref{F:RESULT_VOL1_RANDOM} and~\ref{F:RESULT_VOL2_RANDOM} ) -- such outcome from these models is known in the active learning community~\cite{iwal09}. 


The discriminative and generative views of the dataset utilized in the proposed method attempts to rectify each others mistakes through the queries presented. In Figure~\ref{F:ANALYZE}\subref{F:ERROR_QUERY}, we show the number of misclassified examples in each 100 queries generated by the RF classifier (blue) and the label propagation method (red) generated during one of the trials of the proposed method. Figure~\ref{F:ANALYZE}\subref{F:ERROR_MUTUAL} displays the number of mutually exclusive errors (label propagation correct but RF incorrect and vice versa) on all the unlabeled examples ($E_u$) against the number of iterations. As these plots show, both the classifier and the label propagation techniques initially produce mutually exclusive errors. Over time, these errors are corrected and their respective models are updated accordingly. The steady increase in predictor accuracy (on all the unlabeled examples) plotted in Figure~\ref{F:ANALYZE}\subref{F:ACCURACY}  suggests that this interaction also reduces the common errors made by these two views.

As Figure~\ref{F:ANALYZE}\subref{F:ERROR_QUERY} indicates, both the generative and discriminative views appear to correctly classify the query samples once sufficient number of queries have been presented. Continuing the query process further would generate ambiguous or uncertain samples with confidences near $0$. Intuitively, these samples would improve the margins of the interactive learner. We utilize this empirical observation to devise a \textbf{stopping criterion} for the proposed algorithm: the interactive process is repeated until the query set error reaches zero and then is continued for approximately 500 more examples before terminating.
\subsubsection{Performance of Global~\cite{andres12} method}
So far, we have presented the segmentation performances of the agglomerative method of~\cite{supple14context} on two FIBSEM volumes. Tables~\ref{T:RESULT_GLOBAL_FIB1} and~\ref{T:RESULT_GLOBAL_FIB2} report the results produced by the Global method~\cite{andres12} with boundary predictor learned on full groundtruth (Full GT) data and by the proposed method on the same two FIBSEM volumes. The error values indicate that the proposed method is able to estimate a superpixel boundary classifier as accurate as that leanred from exhaustive groundtruth for Global method~\cite{andres12} as well.
\begin{table}
\begin{center}
\caption{\scriptsize SplitVI average and deviation of Global method~\cite{andres12} on FIBSEM Test vol 1}
\label{T:RESULT_GLOBAL_FIB1}
\begin{tabular}{ | c | c | c | c | c | }
  \hline
  Algorithm & \multicolumn{2}{|c|}{split-VI, Trn set 1}& \multicolumn{2}{|c|}{split-VI, Trn set 2}\\
  \hline
   & false merge & false split & false merge & false split \\
 \hline 
  Full GT  & $0.0824 \pm 0.0035$ & $1.09 \pm 0.0016$ & $0.0814 \pm 0.0005$ & $1.1018 \pm 0.0049$  \\
  Proposed  & $0.0817 \pm 0.0088$ & $1.087 \pm 0.0045$ & $0.0897 \pm 0.0047$ & $1.104 \pm 0.0076$  \\
\hline
\end{tabular}
\end{center}
\end{table}
\begin{table}
\begin{center}
\caption{\scriptsize SplitVI average and deviation of Global method~\cite{andres12} on FIBSEM Test vol 2}
\label{T:RESULT_GLOBAL_FIB2}
\begin{tabular}{ | c | c | c | c | c | }
  \hline
  Algorithm & \multicolumn{2}{|c|}{split-VI, Trn set 1}& \multicolumn{2}{|c|}{split-VI, Trn set 2}\\
  \hline
   & false merge & false split & false merge & false split \\
 \hline
  Full GT  & $0.0348 \pm 0.001$ & $0.86874 \pm 0.0016$ & $0.036 \pm 0.0012$ & $0.858 \pm 0.00265$  \\
  Proposed  & $0.0374 \pm 0.0026$ & $0.8641 \pm 0.004$ & $0.0391 \pm 0.0022$ & $0.871 \pm 0.009$  \\
\hline
\end{tabular}
\end{center}
\end{table}
\subsection{ssTEM data} 
\begin{figure}[t]
\begin{center}
\subfigure[\scriptsize Proposed]{
\includegraphics[width=0.35\columnwidth, height=0.3\columnwidth]{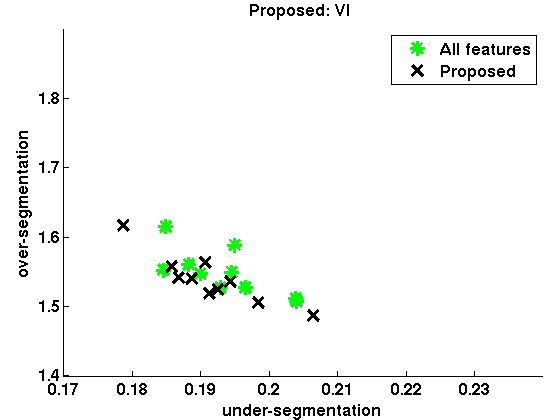}\label{F:PROPOSED_VI_TEM}}
\subfigure[\scriptsize Co-training~\cite{zhu09book}]{
\includegraphics[width=0.35\columnwidth, height=0.3\columnwidth]{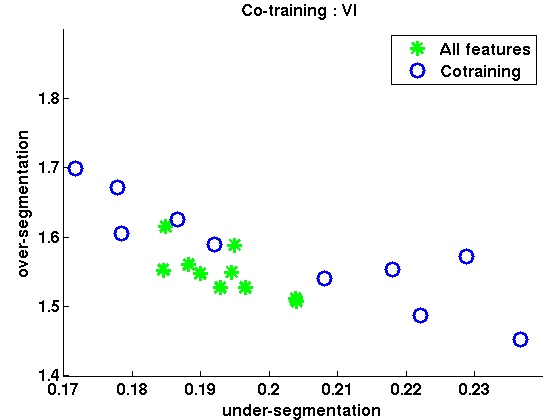}\label{F:COTRAIN_VI_TEM}}
\subfigure[\scriptsize Random]{
\includegraphics[width=0.35\columnwidth, height=0.3\columnwidth]{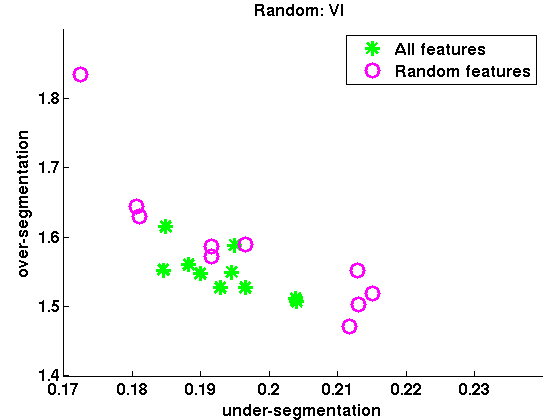}\label{F:RANDOM_VI_TEM}}
\subfigure[\scriptsize Random]{
\includegraphics[width=0.35\columnwidth, height=0.3\columnwidth]{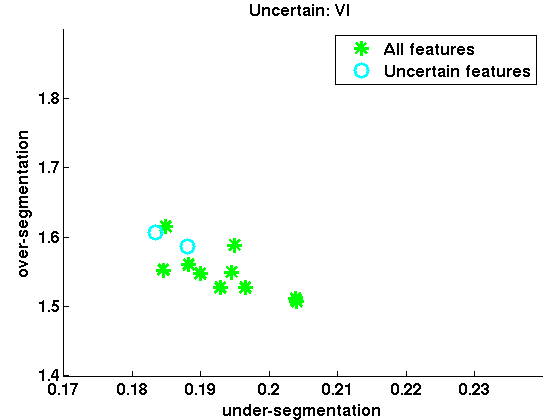}\label{F:UNCERTAIN_VI_TEM}}
\end{center}
\caption{\scriptsize Segmentation error (split-VI) on TEM data.}
\label{F:TEM}
\end{figure}

The proposed algorithm has also been tested on 2D ssTEM  images. We have trained the superpixel boundary predictor on 15 $500 \times 500$ images following the same procedure as used in FIBSEM data and applied on 15 $1000 \times 1000$ pixel images to segment the images (in 2D) in a context-aware fashion. The total number of queries used was $6000$ out of approximately $41000$ samples ($<15\%$). The split-VI errors produced by the predictors learned using the proposed method, random queries and co-training method in 10 trials are shown in Figure~\ref{F:TEM}. In this dataset too, the proposed method estimated a predictor producing the same segmentation errors as those resulted by the ones learned on full groundtruth.

\begin{table}
\label{T:RESULT_VI_TEM}
\begin{center}
\caption{SplitVI average and deviation on TEM images}
\begin{tabular}{ | c | c | c | }
  \hline
  Algorithm & \multicolumn{2}{|c|}{split-VI, Trn set 1}\\
  \hline
   & false merge & false split \\
 \hline
 
  All  & $0.1934 \pm 0.0068$ & $1.5487 \pm 0.0335$  \\
  Proposed  & $0.1943 \pm 0.008235$ & $1.5229 \pm 0.0276$\\
  co-train  & $0.202 \pm 0.0236$ & $1.5796 \pm 0.07617$ \\
  uncertain & $0.1719 \pm 0.01285$ & $1.6843 \pm 0.0887$  \\
  random & $0.1987 \pm 0.0135$ & $1.5658 \pm 0.05952$ \\
\hline
\end{tabular}
\end{center}
\end{table}

\section{Conclusion}
This paper presents an algorithm to train a superpixel boundary predictor from a few of all training examples. The predictors estimated by the the proposed method are shown to be as robust and accurate as those learned from complete groundtruth for segmentation purposes. Such a training algorithm will expedite learning tools for EM segmentation considerably and pave the way for practical semi-automatic segmentation systems for large volumes with diverse region characteristics.

\vspace{0.5cm}
{
\noindent\textbf{Acknowledgement}: The authors are grateful to Stuart Berg of Janelia Farm Research for his support in software development.
}

{
\bibliography{paper768}
\bibliographystyle{splncs}
}

\end{document}